\documentclass[10pt,twocolumn,letterpaper]{article}

\usepackage{cvpr}
\usepackage{times}
\usepackage{epsfig}
\usepackage{graphicx}
\usepackage{amsmath}
\usepackage{amssymb}

\usepackage{multirow}
\usepackage{makecell}

\usepackage[noend]{algpseudocode}
\usepackage{algorithmicx,algorithm}

\usepackage[pagebackref=true,breaklinks=true,letterpaper=true,colorlinks,bookmarks=false]{hyperref}

\cvprfinalcopy 

\def\cvprPaperID{7578} 

\ifcvprfinal\fi
\begin{document}

\title{Exploiting Operation Importance for Differentiable Neural Architecture Search}

\author{Xukai Xie\\
Tianjin university\\
{\tt\small xkxie@tju.edu.cn}
\and
Yuan Zhou\textsuperscript{*}\\
Tianjin university\\
{\tt\small zhouyuan@tju.edu.cn}
\and
Sun-Yuan Kung\\
Princeton University\\
{\tt\small kung@princeton.edu}
}

\maketitle

\begin{abstract}
    Recently, differentiable neural architecture search methods significantly reduce the search cost by constructing a super network and relax the architecture representation by assigning architecture weights to the candidate operations. All the existing methods determine the importance of each operation directly by architecture weights. However, architecture weights cannot accurately reflect the importance of each operation; that is, the operation with the highest weight might not related to the best performance. To alleviate this deficiency, we propose a simple yet effective solution to neural architecture search, termed as exploiting operation importance for effective neural architecture search (EoiNAS), in which a new indicator is proposed to fully exploit the operation importance and guide the model search. Based on this new indicator, we propose a gradual operation pruning strategy to further improve the search efficiency and accuracy. Experimental results have demonstrated the effectiveness of the proposed method. Specifically, we achieve an error rate of 2.50\% on CIFAR-10, which significantly outperforms state-of-the-art methods. When transferred to ImageNet, it achieves the top-1 error of 25.6\%, comparable to the state-of-the-art performance under the mobile setting.
\end{abstract}

\section{Introduction}

Designing reasonable network architecture for specific problems is a challenging task. Better designed network architectures usually lead to significant performance improvement. In recent years, neural architecture search (NAS)~\cite{1,2,3,6,13,15,19,34} has demonstrated success in designing neural architectures automatically. Many architectures produced by NAS methods have achieved higher accuracy than those manually designed in tasks such as image classification~\cite{1}, super resolution~\cite{22}, semantic segmentation~\cite{4,7} and object detection~\cite{11}. NAS methods not only boost the model performance, but also liberate human experts from the tedious architecture tweaking work.

\begin{figure}[t]
\begin{center}
\includegraphics[width=0.8\linewidth]{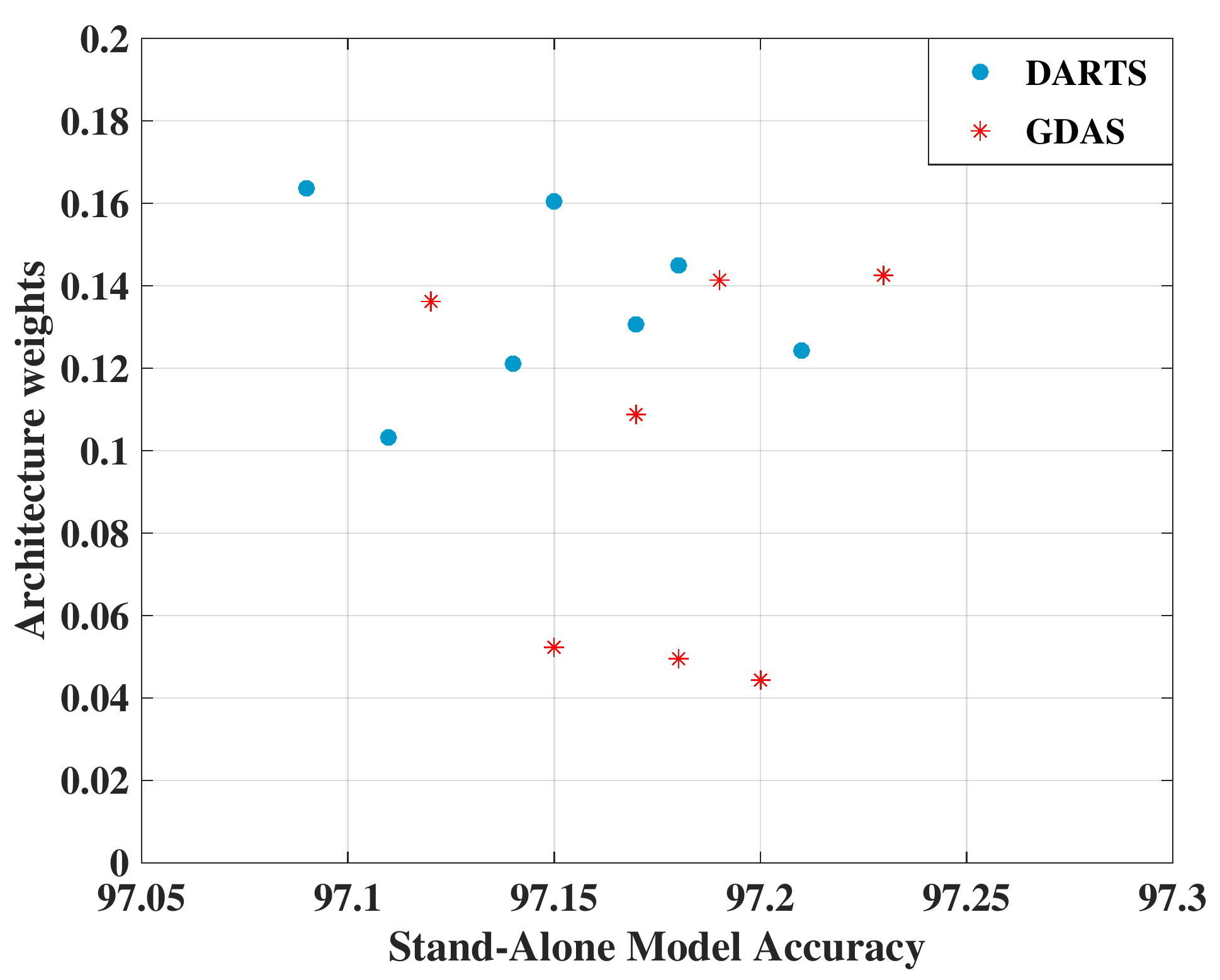}
\end{center}
\caption{Correlation between stand-alone model and learned architecture weights. We replace the selected operation in the first edge of the first cell for the final architecture with all the other candidate operations both in the DARTS~\cite{15} and GDAS~\cite{26}, and fully train them until converge.}
\label{fig:long}
\label{fig:onecol}
\end{figure}

So far, there exist three basic frameworks that have gained a growing interest, \emph{i.e.}, evolutionary algorithm (EA)-based NAS~\cite{14,19,24}, reinforcement learning (RL)-based NAS~\cite{1,2,18}, and gradient-based NAS~\cite{15,25,26}. In both EA-based and RL-based approaches, their searching procedures require the validation accuracy of numerous architecture candidates, which is computationally expensive. For example, the reinforcement learning method~\cite{1,2} trains and evaluates more than 20,000 neural networks across 500 GPUs over 4 days. These approaches use a large amount of computational resources, which is inefficient and unaffordable.

\begin{figure*}[t]
\begin{center}
\includegraphics[width=1.0\linewidth]{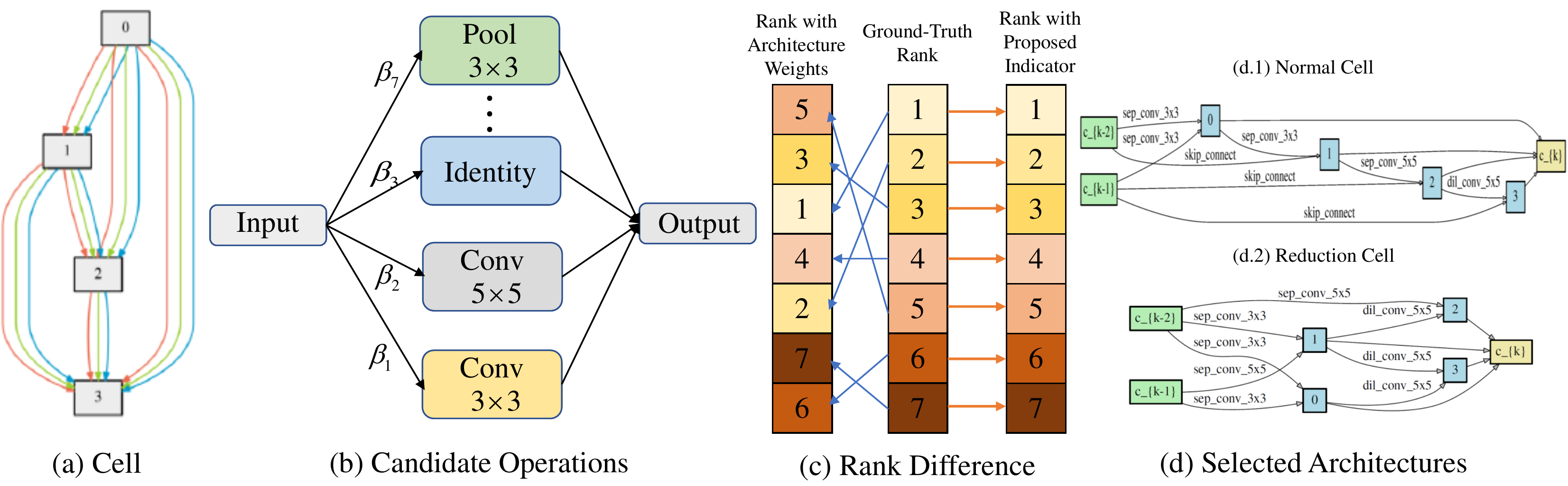}
\end{center}
\caption{Illustration of the NAS procedure. (a) Cell. (b) Candidate operations. (c) Rank Difference: we visualize the corresponding importance of each operation by the colors and numbers; a change in ranking occurs between architecture weight ranking and the true one. (d) Selected architectures.}
\label{fig:long}
\label{fig:onecol}
\end{figure*}

To eliminate such deficiency, gradient-based NAS methods~\cite{15,25,26,33} such as DARTS~\cite{15} and GDAS~\cite{26} are recently presented. They construct a super network and relax the architecture representation by assigning continuous weights to the candidate operations. In DARTS, a computation cell is searched as the building block of the final architecture and each cell is represented as a directed acyclic graph (DAG) consisting of an ordered sequence of \emph{N} nodes. Then the concrete search space is relaxed into a continuous one, so that network and architecture parameters can be well-optimized by gradient descent. It achieved comparable performance to EA-based~\cite{14} and RL-based~\cite{1} methods while only requiring a search cost of a few GPU-days. In order to further accelerate the searching procedure, GDAS~\cite{26} samples one sub-graph according to the architecture weights in a differentiable way at each training iteration.

Existing methods select the candidate operations based on their architecture weights to derive the target architecture. Stand-alone models are constructed to generate weights for all possible architectures in the search space. However, architecture weights cannot accurately reflect the importance of each operation. To illustrate this issue, the obtained accuracy of stand-alone model is compared with the corresponding architecture weights. Their correlation is plotted in Figure 1. We can see that the operation with the highest architecture weight dose not achieves the best accuracy. Furthermore, the architecture weights of candidate operations are often close to each other; in this case, it is difficult to decide which candidate operation is the optimal one. Figure 2 illustrates the procedure of NAS.

Given the limitation of architecture weights, it is natural to ask the question: will we be able to improve architecture search performance if we apply a more effective indicator to guide the model search? To this end, we propose a simple yet effective solution to neural architecture search, termed as exploiting operation importance for effective neural architecture search (EoiNAS). The main idea of our method has two parts:

1) It is well-recognized that operation A is better than operation B if A has fewer training epochs and higher validation accuracy during the search process. According to this criterion, \emph{a new indicator is proposed to fully exploit the operation importance and guide the model search}. Training iterations and validation accuracy for each operation can be recorded in the search space.

2) Based on this new indicator, \emph{we propose a gradual operation pruning strategy to further improve the search efficiency and accuracy}. We denote the training of every k epochs as a step. In each step, we prune the most inferior operation according to the new indicator. This process continues until only one operation remains; this operation can be regarded as the best operation to derive the final architecture. Owing to the gradual operation pruning strategy, our super network exhibits fast convergence.

The effectiveness of EoiNAS is verified on the standard vision setting, i.e., searching on CIFAR-10, and evaluating on both CIFAR-10/100 and ImageNet datasets. We achieve state-of-the-art performance of 2.50\% test error on CIFAR-10 using 3.4M parameters. When transferred to ImageNet, it achieves top-1/5 errors of 25.6\%/8.3\% respectively, comparable to the state-of-the-art performance under the mobile setting.

The remainder of this paper is organized as follows: In Section 2, we review the related work of recent neural architecture search algorithms and describe our search method in Section 3. After experiments are shown in Section 4, we conclude this paper in Section 5.

\section{Related Work}

With the rapid development of deep learning, significant gain in performance has been brought to a wide range of computer vision problems, most of which owed to manually designed network architectures~\cite{8, 10, 12, 21, 23, 27}. Recently, a new research field named neural architecture search (NAS)~\cite{1, 2, 3, 6, 13} has been attracting increasing attentions. The goal is to find automatic ways of designing neural architectures to replace conventional handcrafted ones. According to the heuristics to explore the large architecture space, existing NAS approaches can be roughly divided into three categories, namely, evolutionary algorithm-based approaches~\cite{14, 19, 24}, reinforcement learning-based approaches~\cite{1, 2, 18} and gradient-based approaches~\cite{15, 25, 26}.

\textbf{Reinforcement learning based NAS}. A reinforcement learning based approach has been proposed by Zoph et al.~\cite{1,2} for neural architecture search. They use a recurrent network as a controller to generate the model description of a child neural network designated for a given task. The resulted architecture (NASNet) improved over the existing hand-crafted network models at its time.

\textbf{Evolutionary algorithm-based NAS}. An alternative search technique has been proposed by Real et al.~\cite{19} where an evolutionary (genetic) algorithm has been used to find a neural architecture tailored for a given task. The evolved neural network (AmoebaNet), further improved the performance over NASNet. Although these works achieved state-of-the-art results on various classification tasks, their main disadvantage is the large amount of computational resources they demand.

\textbf{Gradient-based NAS}. Contrary to treating architecture search as a black-box optimization problem, gradient based neural architecture search methods~\cite{15,25,26} utilized the gradient obtained in the training process to optimize neural architecture. DARTS~\cite{15} relaxed the search space to be continuous, so that the architecture can be optimized with respect to its validation set performance by gradient descent. Therefore, gradient-based approaches successfully accelerate the architecture search procedure, only several GPU days are required. Because DARTS optimized the entire super network during the search process, it may suffer from discrepancy between the continuous architecture encoding and the derived discrete architecture. GDAS~\cite{26} suggested an alternative method to alleviate this discrepancy. GDAS approaches the search problem as sampling from a distribution of architectures, where the distribution itself is learned in a continuous way. The distribution is expressed via slack softened one-hot variables that multiply the operations and make the sampling procedure differentiable. SNAS~\cite{25} applied a similar technique to constrain the architecture parameters to be one-hot to tackle the inconsistency in optimizing objectives between search and evaluation scenarios. In order to bridge the depth gap between search and evaluation scenarios, PDARTS~\cite{33} divide the search process into multiple stages and progressively increase the network depth at the end of each stage. In addition, MdeNAS~\cite{32} propose a multinomial distribution learning method for extremely effective NAS, which considers the search space as a joint multinomial distribution and the distribution is optimized to have high expectation of the performance.

\begin{figure}[t]
\begin{center}
\includegraphics[width=0.8\linewidth]{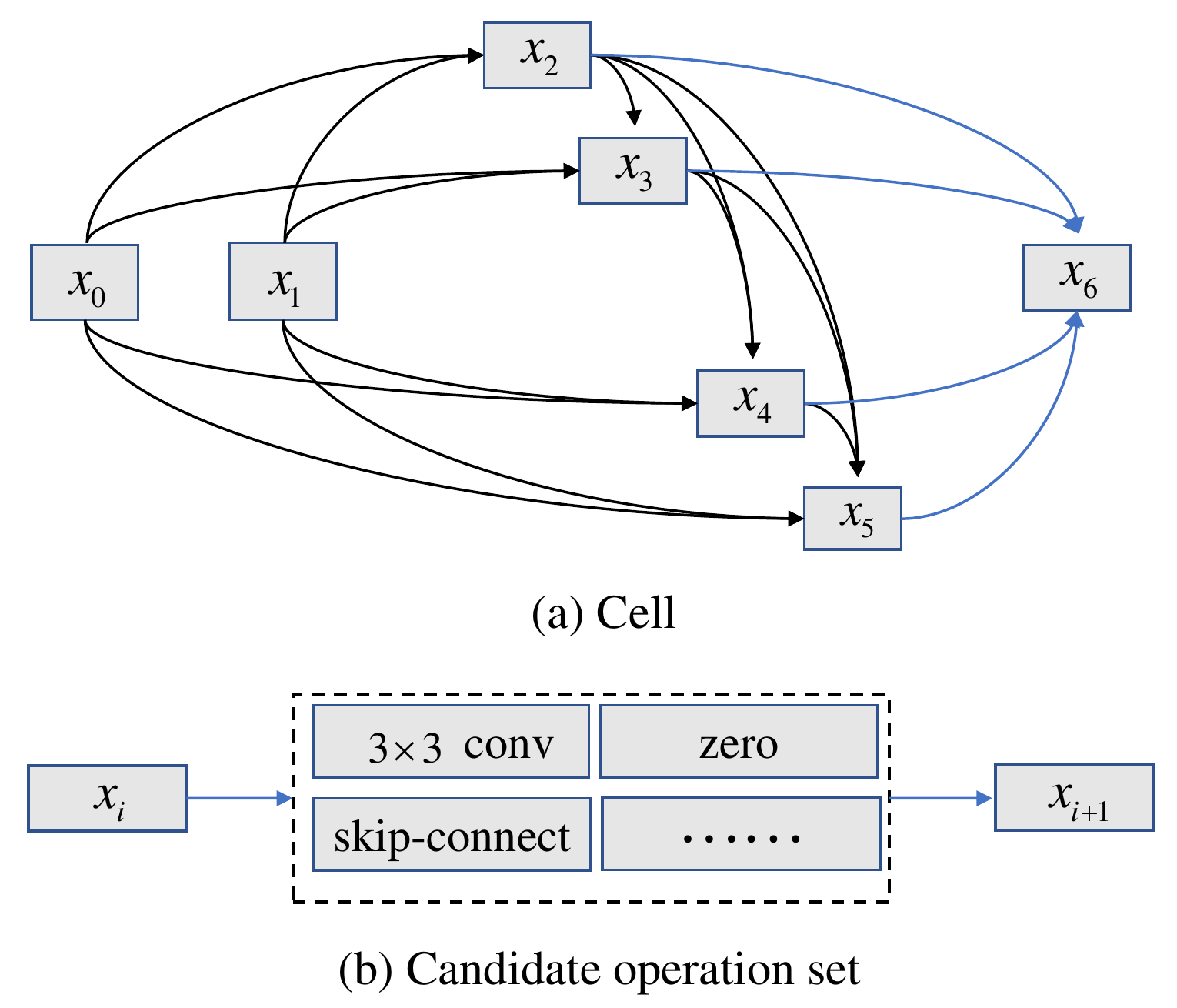}
\end{center}
\caption{ Search space. (a) A cell contains 7 nodes, two input nodes, four intermediate nodes that apply sampled operations on the input nodes and upper nodes, and an output node that concatenates the outputs of the four intermediate nodes. (b) The edge between two nodes denotes a possible operation sampled according to the discrete probability distribution $\Gamma$ in the search space.}
\label{fig:long}
\label{fig:onecol}
\end{figure}

\section{Methodology}

We first describe our search space and continuous relaxation in general form in Section 3.1, where the computation procedure for an architecture is represented as a directed acyclic graph. We then propose a new indicator to fully exploit the importance of each operation in Section 3.2. Finally, we design an gradual operation pruning strategy to make the super network exhibit fast convergence and high training accuracy in Section 3.3.

\subsection{Search Space and Continuous Relaxation}

In this work, we leverage GDAS~\cite{26} as our baseline framework. Our goal is to search a robust cell and apply it to a network of \emph{L} cells. As shown in Figure 3, a cell is defined as a directed acyclic graph (DAG) of \emph{N} nodes, $\{ x_0, x_2, \cdots, x_{N-1} \}$, where each node is a network layer, \emph{i.e.}, performing a specific mathematical function. We denote the operation space as $ O $ , in which each element represents a candidate operation $o(\cdot)$. An edge $f_{i,j}$ represents the information flow connecting node $x_{i}$ and $x_{j}$, which consists of a set of operations weighted by the architecture weights $\beta_{i,j}$, and is thus formulated as:

\begin{equation}
\label{1}
f_{i,j}(x_i)= \sum\limits_{o\in O} \beta_{i,j}^o o(x_i)
\end{equation}

\begin{equation}
\label{2}
\mathrm{s.t.}\hspace{2mm} \beta_{i,j}^o= \frac{exp(\alpha_{i,j}^o)}{\sum_{o^{'}\in O} exp(\alpha_{i,j}^{o^{'}})}
\end{equation}
where $\alpha_{i,j}^o$ is the \emph{o-th} element of an \emph{O}-dimensional learnable vector $\alpha_{i,j}\in\mathbb{R}^{O}$, and $\beta_{i,j}$ encodes the sampling distribution of the function between node $x_{i}$ and $x_{j}$, as we will discuss below. Intuitively, a well learned $\beta=\{\beta_{i,j}^o\}$ could represent the relative importance of the operation \emph{o} for transforming the feature map $x_{i}$. Similar to GDAS, between node $x_{i}$ and $x_{j}$, we sample one operation from $ O $ according to a discrete probability distribution $\Gamma_{i,j}$ which is characterized by Eq. (2). During the search, we calculate each node in a cell as:

\begin{equation}
\label{3}
x_{j}= \sum_{i\textless j} f_{i,j}(x_{i})
\end{equation}
where $f_{i,j}$ is sampled from $\Gamma_{i,j}$.

Since the operation $f_{i,j}$ is sampled from a discrete probability distribution, we cannot back-propagate gradients to optimize $\alpha_{i,j}$. To allow back-propagation, we use the Gumbel-Max trick~\cite{27,30} and softmax function~\cite{28} to re-formulate Eq. (3) to Eq. (4), which provides an efficient way to draw samples from a discrete probability distribution in a differentiable way.

\begin{equation}
\label{4}
x_{j}= \sum_{i=1}^{j-1} \sum_{o=1}^O h_{i,j}^o f_{i,j}^o (x_{i};W_{i,j}^o)
\end{equation}

\begin{equation}
\label{6}
\mathrm{s.t.}\hspace{2mm} h_{i,j}^o = \frac{exp((\alpha_{i,j}^o+\tau_o)/T)}{\sum_{o^{'}=1}^O exp((\alpha_{i,j}^{o^{'}}+\tau_{o^{'}})/T)}
\end{equation}
Here $\tau_o$ are i.i.d samples drawn from Gumbel (0,1); $f_{i,j}^o$  indicates the \emph{o}-th function in \emph{O}; $h_{i,j}^o$ is the \emph{o}-th element of $h_{i,j}$; $W_{i,j}^o$ is the weight of $f_{i,j}^o$ for the transformation function between node $x_{i}$ and $x_{j}$; \emph{T} is the temperature parameter~\cite{27}, which controlls the Gumbel-Softmax distribution. As the parameter \emph{T} approaches zero, the Gumbel-Softmax distribution becomes equivalent to the discrete probability distribution. The temperature parameter is annealed from 5.0 to 0.0 during our search.

Our candidate operation set $O$ contains the following 8 operations: (1) identity, (2) zero, (3) $3\times 3$ separable convolutions, (4) $3\times 3$ dilated separable convolutions, (5) $5\times 5$ separable convolutions, (6) $5\times 5$ dilated separable convolutions, (7) $3\times 3$ average pooling, (8) $3\times 3$ max pooling. We search for two kinds of cells, \emph{i.e.}, the normal cell and the reduction cell. When searching the normal cell, each operation in $O$ has the stride of 1. For the reduction cell, the stride of operations on 2 input nodes is 2. Once we discover the best normal cell and reduction cell, we stack copies of these best cells to make up a neural network.

\begin{figure}[t]
\begin{center}
\includegraphics[width=0.75\linewidth]{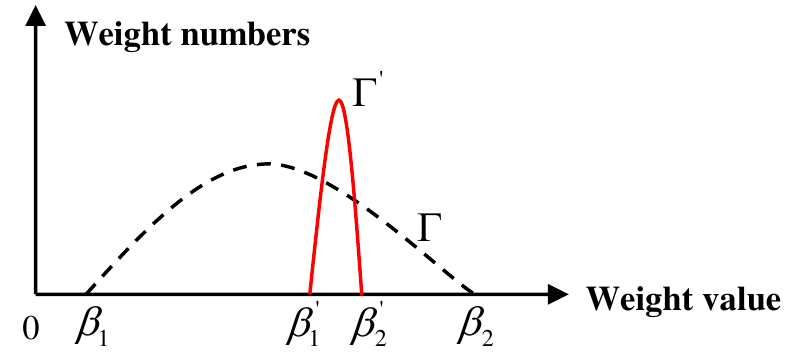}
\end{center}
\caption{Ideal and reality cases of the weight deviation. The dark dashed curve indicates the ideal weight distribution, and the red solid curve denotes the weight distribution might occur in real cases. The deviation of architecture weights should be large enough, so that we can clearly judge which operation is more important.}
\label{fig:long}
\label{fig:onecol}
\end{figure}

\begin{figure}[h]
\begin{center}
\includegraphics[width=0.72\linewidth]{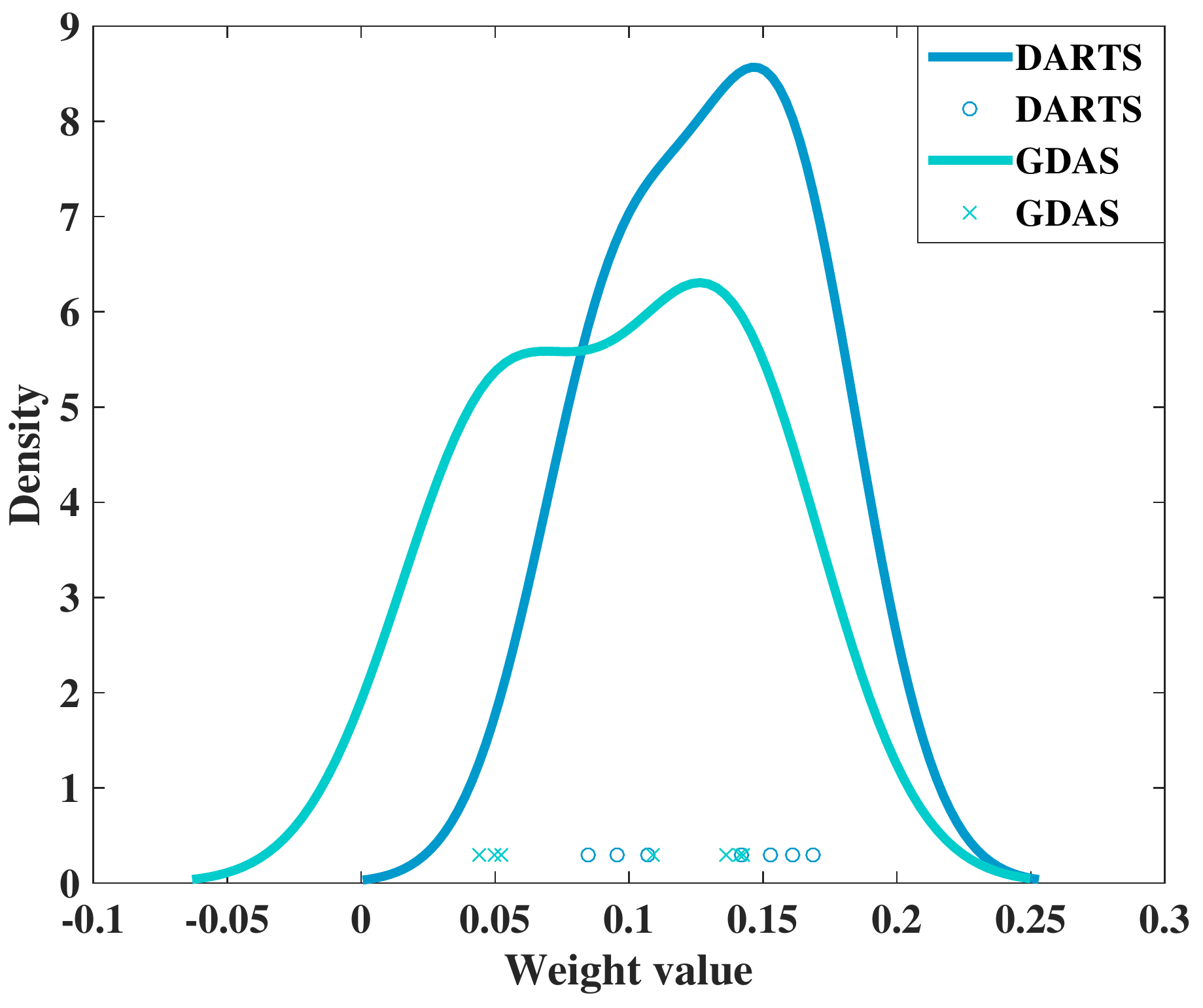}
\end{center}
\caption{Distribution of architecture weights. The small circles, '$\times$' and solid curves denote each operation weight and Kernel Distribution Estimate (KDE) of the weight distribution respectively. Architecture weights are distributed too densely, which makes it difficult to distinguish the important operations from the others.}
\label{fig:long}
\label{fig:onecol}
\end{figure}

\subsection{Operation Importance Indicator}

\textbf{Architecture Weights Deviation.} In previous algorithms, operation importance is ranked by the architecture weights $\beta$, which is supposed to represent the relative importance of a candidate operation verse the others. When the search process is over, they select the most important operation and prune other inferior operations according to the value of the architecture weights.

However, architecture weights cannot accurately reflect the importance of each operation. As shown in Figure 4, the dark dashed curve and the red solid curve indicates the architecture distribution in ideal and real cases respectively. In ideal cases, the deviation of architecture weights is large enough, so that we can clearly judge which operation is more important. However, this requirement may not always hold and it might lead to unexpected results. In Figure 5, statistical information collected from the cell of DARTS and GDAS demonstrates this analysis. The small circles and '$\times$' show each observation in this weight distribution, and the solid curves denote the Kernel Distribution Estimate (KDE)~\cite{38}, which is a non-parametric way to estimate the probability density function of a random variable. As shown in Figure 5, there is a large quantity of operations whose architecture weights are distributed on a small interval, which makes it difficult to distinguish the important operations from the others. Figure 2 (c) also illustrates this issue: a change in ranking occurs between architecture weight ranking and the true one.

\textbf{The Proposed Indicator.} It is well-recognized that operation A is better than operation B if A has fewer training epochs and higher validation accuracy during the search process. Therefore, For each operation,  the ratio of training iterations and validation accuracy can be used to determine the operation importance. This ratio is represented by
\begin{equation}
\label{8}
C_{i,j}^o = \frac{C_{i,j}^{a^o}}{C_{i,j}^{e^o}}
\end{equation}
where $C_{i,j}^{a^o}$ and $C_{i,j}^{e^o}$ is the validation accuracy and training iterations of each operation on each edge respectively.
The value of accuracy parameters might also close to each other, which will affect importance judgement; in this case, we consider the operation with higher architecture weights will be more important. Therefore, we combine accuracy parameters with architecture weights to obtain an effective indicator \emph{I} as Eq. (7), which can fully exploit the importance of each operation.
\begin{equation}
\label{7}
I_{i,j}^o = \beta_{i,j}^o + \lambda C_{i,j}^o
\end{equation}
where $\beta_{i,j}^o$ is the architecture weights of the \emph{o}-th operation between node $x_{i}$ and $x_{j}$, $\lambda$ is a parameter to control the balance between the two parts, which is set to 0.5 in this work.

Compared to previous methods~\cite{15,26} that judge the operation importance directly by architecture weights, our proposed indicator can effectively reflect the operation importance, which can help to select the optimal operation, so as to achieve the highest accuracy. Apply this effective indicator can be able to improve architecture search performance significantly.

Based on this new indicator \emph{I}, gradual operation pruning strategy is proposed during the search process to further improve the search efficiency and accuracy, as we will discuss next.

\begin{table*}[t!]
\centering
\setlength{\tabcolsep}{5.3pt}
\begin{tabular}{| c | c | c | c | c | c | c | c |} \hline\hline

\multirow{2}{*}{\textbf{Type}} & \multirow{2}{*}{\textbf{Architecture}} & \multirow{2}{*}{\textbf{GPUs}} & \textbf{Times} & \textbf{Params}  & \multicolumn{2}{c|}{\textbf{Test Error}} & \multirow{2}{*}{\textbf{Search Method}}\\\cline{6-7}
    & & & (days) & (million) & C10(\%) & C100(\%) & \\ \hline
\multirow{3}{*}{\makecell{Human\\expert}}
    &  ResNet + CutOut~\cite{8}       & $-$ & $-$   & 1.7  & 4.61       & 22.10   & manual \\
    &  DenseNet~\cite{10}              & $-$ & $-$   & 25.6 & 3.46       & 17.18   & manual \\
    &  SENet~\cite{31}                  & $-$ & $-$   & 11.2 & 4.05       & $-$     & manual \\
      \hline\hline
\multirow{19}{*}{\makecell{Neural\\architecture\\search}}
    &  MetaQNN~\cite{36}                & 10  & 8-10  & 11.2   & 6.92      &  27.14   & RL \\
    &  NAS~\cite{1}                    & 800 & 21-28 & 7.1    & 4.47      &  $-$  & RL \\
    &  NASNet-A~\cite{2}               & 450 & 3-4   & 3.3    & 3.41      &  $-$  & RL \\
    &  NASNet-A + CutOut~\cite{2}      & 450 & 3-4   & 3.3    & 2.65      &  $-$  & RL \\
    &  ENAS~\cite{18}                   & 1   & 0.45  & 4.6    & 3.54      &  19.43  & RL \\
    &  ENAS + CutOut~\cite{18}          & 1   & 0.45  & 4.6    & 2.89      &  $-$  & RL \\
    &  AmoebaNet-A + CutOut~\cite{19}    & 450 & 7.0     & 3.2    & 3.34 &  18.93  & evolution  \\
    &  AmoebaNet-B + CutOut~\cite{19}    & 450 & 7.0     & 2.8    & 2.55 &  $-$  & evolution  \\
    &  Hierarchical NAS~\cite{14}       & 200 & 1.5   & 61.3   & 3.63      &  $-$  & evolution  \\
    &  Progressive NAS~\cite{13}        & 100 & 1.5   & 3.2    & 3.63      &  19.53  & SMBO \\
    &  DARTS (1st) + CutOut~\cite{15}   & 1   & 0.38  & 3.3    & 3.00      &  17.76  & gradient-based  \\
    &  DARTS (2nd) + CutOut~\cite{15}   & 1   & 1.0     & 3.4    & 2.82      &  17.54  & gradient-based  \\
    &  SNAS + CutOut~\cite{25}          & 1   & 1.5   & 2.9    & 2.98      &  $-$  & gradient-based  \\
    &  GDAS~\cite{26}                   & 1   & 1.0     & 3.4    & 3.87      &  19.68  & gradient-based  \\
    &  GDAS + CutOut~\cite{26}          & 1   & 1.0     & 3.4    & 2.93      &  18.38  & gradient-based  \\
    &  MdeNAS + CutOut~\cite{32}        & 1   & 0.16  & 3.61   & 2.55      &  $-$  & MDL  \\
    &  Random Search + CutOut~\cite{15} & 1   & 4.0     & 3.2    & 3.29      &  $-$  & random  \\
    \cline{2-8}
    &  EoiNAS                & 1   & 0.6  & 3.4    & 3.42      &  18.4  & gradient-based  \\
    &  EoiNAS + CutOut       & 1   & 0.6  & 3.4    & 2.50      &  17.3  & gradient-based  \\
    \hline
    \hline
\end{tabular}
\vspace{2mm}
\caption{
Classification errors of EOINAS and benchmarks on CIFAR-10 and CIFAR-100.
}
\vspace{-2mm}
\label{table:CIFAR}
\end{table*}

\subsection{Gradual Operation Pruning Strategy}

In existing methods, all candidate operations are always kept during the search process and unimportant operations are removed directly by the architecture weights until the search is over to derive the final architecture. However, for some unimportant operations, we do not need to waste time and computation resources to sample and train.

Therefore, we propose a gradual operation pruning strategy to further improve the search efficiency and accuracy. We denote the training of every \emph{k} epochs as a step. In a step, we prune the most inferior operation according to the new indicator. In the next step, we judge the most inferior operation in the remaining operations and prune it. This process continues until only one operation remains; this operation can be regarded as the best operation to derive the final architecture. Owing to the gradual operation pruning strategy, our super network exhibits fast convergence.

Our definite searching algorithm is presented in Algorithm 1. At the initialization of the search process, we perform gradient-descent based optimization over the network parameters in the first 20 epochs. It helps obtaining balanced architecture weights between parameterized operations (\emph{e.g}. convolution operation) and non-parameterized operations (\emph{e.g}. skip-connect operation). Then, we perform a gradient-descent based optimization for the architecture parameters $\alpha$ and network parameters $w$ in an alternating manner. Specifically, we optimize the operation weights by descending $\bigtriangledown_wL_{train}(w,\alpha;T)$ on the training set, and optimize the architecture parameters by descending $\bigtriangledown_{\alpha}L_{val}(w,\alpha;T)$ on the validation set. An operation will be pruned after 20 epochs if its corresponding operation importance indicator $I_{i,j}^o$, which is updated along training iterations, is the lowest. When the search procedure is finished, we decode the discrete cell architecture by first retaining the two strongest predecessors for each node (with the strength from node $x_{i}$ and $x_{j}$, being $max_{o, O^o\neq zero}I_{i,j}^o$), and then choose the most likely operation by taking the argmax.

\begin{algorithm}[t]
\caption{Efficient Neural Architecture Search}
\hspace*{0.02in} {\bf Input:}
Training set: $D_T$; Validation set: $D_V$; \\
\hspace*{0.40in} {}
Operation set: $O$ \\
\hspace*{0.02in} {\bf Init:}
Network parameters: $w$;  Architecture parameters: $\alpha$; \\
\hspace*{0.28in} {}
Validation accuracy: $C^a$; Training iterations: $C^e$; \\
\hspace*{0.28in} {}
Temperature parameter: $T$
\begin{algorithmic}[1]
\While{not converge}
    \State Sample a sub-graph to train according to Eq. (5);
    \State Update $w$  by $\bigtriangledown_wL_{train}(w,\alpha;T)$ on $D_T$;
    \State Update $\alpha$ by $\bigtriangledown_{\alpha}L_{val}(w,\alpha;T)$ on $D_V$;
    \State Update $C^e$, $C^a$;
    \If{epoch\textgreater 20 and epoch \% 20 == 0}
        \State Calculate operation importance $I$ by Eq. (7);
        \State $O=O\backslash\{O_{argmin_{o}I^o} \}$
    \EndIf
\EndWhile
\State \hspace*{0.18in} {\bf end if}\\
{\bf end while}
\State Derive the final architecture based on the indicator \emph{I};
\State Optimize the architecture on the training set
\end{algorithmic}
\end{algorithm}

\section{Experiments}

\subsection{Datasets}

We conduct experiments on three popular image classification datasets, including CIFAR-10, CIFAR-100~\cite{29} and ImageNet~\cite{20}. Architecture search is performed on CIFAR-10, and the discovered architectures are evaluated on all three datasets.

Both CIFAR-10 and CIFAR-100 have 50K training and 10K testing RGB images with a fixed spatial resolution of $32\times 32$. These images are equally distributed over 10 classes and 100 classes in CIFAR-10 and CIFAR-100 respectively. In the architecture search scenario, the training set is equally split into two subsets, one for updating network parameters and the other for updating the architecture parameters. In the evaluation scenario, the standard training/testing split is used.

We use ImageNet to test the transferability of the architectures discovered on CIFAR-10. Specificaly, we use a subset of ImageNet, namely ILSVRC2012, which contains 1,000 object categories and 1.28M training and 50K validation images. Following the conventions~\cite{2,15}, we apply the mobile setting where the input image size is $224\times 224$.

\begin{table*}[t!]
\centering
\setlength{\tabcolsep}{4.5pt}
\begin{tabular}{| c | c | c | c | c | c | c | c | c |} \hline\hline

  \multirow{2}{*}{\textbf{Type}} &  \multirow{2}{*}{\textbf{Architecture}}  & \multirow{2}{*}{\textbf{GPUs}} & \textbf{Times} & \textbf{Params} & \textbf{MAdds} & \multicolumn{2}{c|}{\textbf{Test Error} (\%)} & \multirow{2}{*}{\textbf{Search Method}}\\ \cline{7-8}
    &  & & (days) & (million) & (million) & Top-1 & Top-5 & \\ \hline

 \multirow{4}{*}{\makecell{Human\\expert}}
    &  Inception-v1~\cite{27}                     & $-$  & $-$   &   6.6  & 1448  & 30.2    &   10.1  & manual\\
    &  MobileNet-V2~\cite{16}                     & $-$  & $-$   &   3.4  & 300   & 28.0    &   $-$   & manual\\
    &  MobileNet-V3~\cite{17}                     & $-$  & $-$   &   5.4  & 219   & 24.8    &   $-$   & manual\\
    &  ShuffleNet~\cite{9}                       & $-$  & $-$   &   5.0  & 524   & 26.3    &   $-$   & manual\\
      \hline\hline
\multirow{12}{*}{\makecell{Neural\\architecture\\search}}
    &  NASNet-A~\cite{2}                         & 450  & 3-4   & 5.3   & 564    &   26.0  & 8.4    & RL \\
    &  NASNet-B~\cite{2}                        & 450  & 3-4   & 5.3   & 488    &   27.2  & 8.7    & RL \\
    &  NASNet-C~\cite{2}                        & 450  & 3-4   & 4.9   & 558    &   27.5  & 9.0    & RL \\
    &  AmoebaNet-A~\cite{19}                     & 450  & 7.0     & 5.1   & 555    &   25.5  & 8.0    & evolution\\
    &  AmoebaNet-B~\cite{19}                     & 450  & 7.0     & 5.3   & 555    &   26.0  & 8.5    & evolution\\
    &  AmoebaNet-C~\cite{19}                     & 450  & 7.0     & 6.4   & 570    &   24.3  & 7.6    & evolution\\
    &  Progressive NAS~\cite{13}                 & 100  & 1.5   & 5.1   & 588    &   25.8  & 8.1    & SMBO\\
    &  DARTS (2nd)~\cite{15}                     & 1    & 1.0     & 4.9   & 595    &   26.9  & 9.0    & gradient-based\\
    &  SNAS~\cite{25}                            & 1    & 1.5   & 4.3   & 522    &   27.3  & 9.2    & gradient-based\\
    &  GDAS~\cite{26}                            & 1    & 1.0     & 5.3   & 581    &   26.0  & 8.5    & gradient-based\\
    &  MdeNAS~\cite{32}                          & 1    & 0.16  & 6.1   & 596    &   25.5  & 7.9    & MDL\\
    \cline{2-9}
    & EoiNAS                           & 1    & 0.6   & 5.0   & 570    &   25.6  & 8.3    & gradient-based\\
    \hline\hline
\end{tabular}
\vspace{2mm}
\caption{
 Comparison with the state-of-the-art architectures on ImageNet (mobile setting). All the NAS networks are searched on CIFAR-10, and then directly transferred to ImageNet.
}
\vspace{-2mm}
\label{table:ImageNet}
\end{table*}

\begin{figure}[t]
\begin{center}
\includegraphics[width=1.0\linewidth]{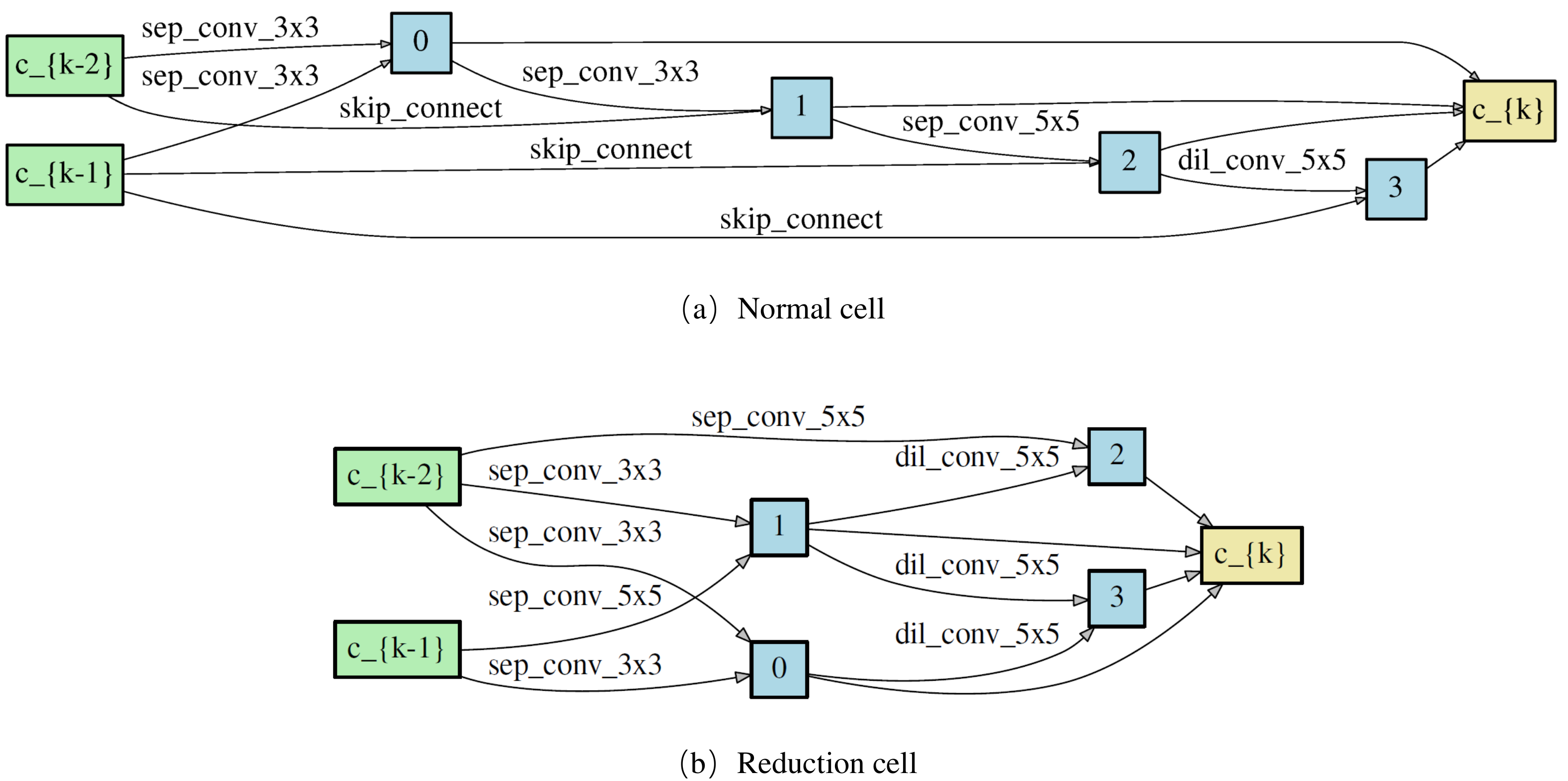}
\end{center}
\caption{Detailed structure of the best cells discovered on CIFAR-10 by our EoiNAS. (a) Normal cell. (b) Reduction cell. The definition of the operations on the edges is in Section 3.1. In the normal cell, the stride of operations on 2 input nodes is 1 and the stride is 2 in the reduction cell.}
\label{fig:long}
\label{fig:onecol}
\end{figure}

\subsection{Implementation Details}

Following the pipeline in GDAS~\cite{26}, our experiments consist of three stages. First, EoiNAS is applied to search for the best normal/reduction cells on CIFAR-10. Then, a larger network is constructed by stacking the learned cells and retrained on both CIFAR-10 and CIFAR-100. The performance of EoiNAS is compared with other state-of-the-art NAS methods. Finally, we transfer the cells learned on CIFAR-10 to ImageNet to evaluate their performance on larger datasets.

\textbf{Network Configrations}. The neural cells for CNN are searched on CIFAR-10 following~\cite{15,25,26}. The candidate function set \emph{O} has 8 different functions as introduced in Section 3.1. By default, we train a small network of 8 cells for 160 epochs in total and set the number of initial channels in the first convolution layer C as 16. Cells located at the 1/3 and 2/3 of the total depth of the network are reduction cells, in which all the operations adjacent to the input nodes are of stride two.

\textbf{Parameter Settings}. For network parameters \emph{w}, we use the SGD optimization. We start with a learning rate of 0.025 and anneal it down to 0.001 following a cosine schedule. We use the momentum of 0.9 and the weight decay of 0.0003. For architecture parameters $\alpha$ ,we use zero initialization which implies equal amount of attention over all possible operations. And we use the Adam optimization~\cite{35}  with the learning rate of 0.0003, momentum (0.5; 0.999) and the weight decay of 0.001. To control the temperature parameter \emph{T} of the Gumbel Softmax in Eq. (5), we use an exponentially decaying schedule. The \emph{T} is initialized as 5 and finally reduced to 0. Following~\cite{26}, we run EoiNAS 4 times with different random seeds and pick the best cell based on its validation performance. This procedure can reduce the high variance of the searched results.

Our EoiNAS takes about 0.6 GPU days to finish the search procedure on a single NVIDIA 1080Ti GPU.  The best cells searched by EoiNAS is shown in Figure 6.

\subsection{Results on CIFAR-10 and CIFAR-100}

For CIFAR, we built a network  with 20 cells and 36 input channels, and trained it by 600 epochs with batch size 128. Cutout regularization~\cite{5} of length 16, drop-path of probability 0.3 and auxiliary towers of weight 0.4~\cite{1} are applied. A standard SGD optimizer with a weight decay of 0.0003 and a momentum of 0.9 is used. The initial learning rate is 0.025, which is decayed to 0 following the cosine rule.

Evaluation results and comparison with state-of-the-art approaches are summarized in Table 1. As demonstrated in Table 1, EoiNAS achieves test errors of 2.50\% and 17.3\% on CIFAR-10 and CIFAR-100, respectively, with a search cost of only 0.6 GPU-days. To obtain the same performance, AmoebaNet~\cite{19} spent four orders of magnitude more computational resources (0.6 GPU-days vs 3150 GPU-days). Our EoiNAS also outperforms GDAS~\cite{26} and SNAS~\cite{25} by a large margin. Notably, architectures discovered by EoiNAS outperform MdeNAS~\cite{32}, the previously most efficient approach, while with fewer parameters. In addition, we compare our method to random search (RS)~\cite{15}, which is considered as a very strong baseline. Note that the accuracy of the model searched by EoiNAS is 0.7\% higher than that of RS.

\subsection{Results on ImageNet}

The ImageNet dataset is used to test the transferability of architectures discovered on CIFAR-10. We adopt the same network configurations as GDAS~\cite{26}, \emph{i.e.}, a network of 14 cells and 48 input channels. The network is trained by 250 epochs with batch size 128 on a single NVIDIA 1080Ti GPU, which takes 12 days with the PyTorch~\cite{37} implementation. The network parameters are optimized using an SGD optimizer with an initial learning rate of 0.1 (decayed linearly after each epoch), a momentum of 0.9 and a weight decay of $3\times 10^{-5}$. Additional enhancements including label smoothing and auxiliary loss tower are applied during training.

Evaluation results and comparison with state-of-the-art approaches are summarized in Table 2. Architecture discovered by EoiNAS outperforms that by GDAS by a large margin in terms of classification accuracy and model size. It demonstrates the transfer capability of the discovered architecture from small dataset to large dataset.

\subsection{Ablation Studies}

In addition, we have conducted a series of ablation studies that validate the importance and effectiveness of the proposed operation importance indicator as well as gradual operation pruning strategy incorporated in the design of EoiNAS.

In Table 3, we show ablation studies on CIFAR-10. GDAS~\cite{26} is our baseline frame work. The GOP means the gradual operation pruning strategy, the OII means the proposed operation importance indicator. All architectures are trained by 600 epochs. As the results show, our super network exhibits fast convergence and high training accuracy owing to the gradual operation pruning strategy. The structure of the best cells discovered on CIFAR-10 is shown in Figure 7. Through prune inferior operations gradually during the search process, we achieve much improvement in performance while using less search times.

Table 3 also demonstrated the effectiveness of the proposed operation importance indicator. The proposed indicator can better judge the importance of each operation and achieve higher accuracy. Such results reveal the necessity of the operation importance indicator.

\begin{table}[h!]
\centering
\setlength{\tabcolsep}{4.5pt}
\begin{tabular}{| c | c | c | c | c | c |} \hline\hline
\multirow{2}{*}{\textbf{Architecture}} &  \multirow{2}{*}{\textbf{GOP}}   & \multirow{2}{*}{\textbf{OII}} & \textbf{Times} & \textbf{Params} & \textbf{Error}\\
& & & (days) &  (million) & (\%) \\\hline
Baseline         &  $\times$             &  $\times$       &  1.0     & 3.4  &  2.93   \\
Baseline+GOP    &  $\checkmark$         &  $\times$       &  0.6   & 3.4  &  2.72   \\
\hline\hline
EoiNAS            &  $\checkmark$         &  $\checkmark$   &  0.6   & 3.4  &  2.50   \\
\hline\hline
\end{tabular}
\vspace{2mm}
\caption{
Ablation studies on CIFAR-10. The GDAS~\cite{26} is our baseline. The GOP means the gradual operation pruning strategy, the OII means the proposed operation importance indicator. All architectures are trained by 600 epochs.}
\vspace{-2mm}
\label{table:WT2}
\end{table}

\subsection{Searched Architecture Analysis}

In differentiable NAS methods, architecture weights is not able to accurately reflect the importance of each operation as discussed in Section 1, because the accuracy of the fully trained stand-alone model and their corresponding architecture weights have low correlation. The proposed operation importance indicator can better decide which operation should be keep on each edge and which edges should be the input of each node, especially for the selection of skip-connect.

The skip-connect operation plays an important role in cell structure. As well studied in~\cite{39,40}, including a reasonable number and location of skip connections would make the gradient flows easier and optimization of deep neural network more stable. Compared the searched results in Figure 6 and Figure 7, architecture discovered by EoiNAS on CIFAR-10 tend to preserve the skip-connect operations in a hierarchical way, which can facilitate  gradient back propagation and make the network have a better convergence

\begin{figure}[t]
\begin{center}
\includegraphics[width=1.0\linewidth]{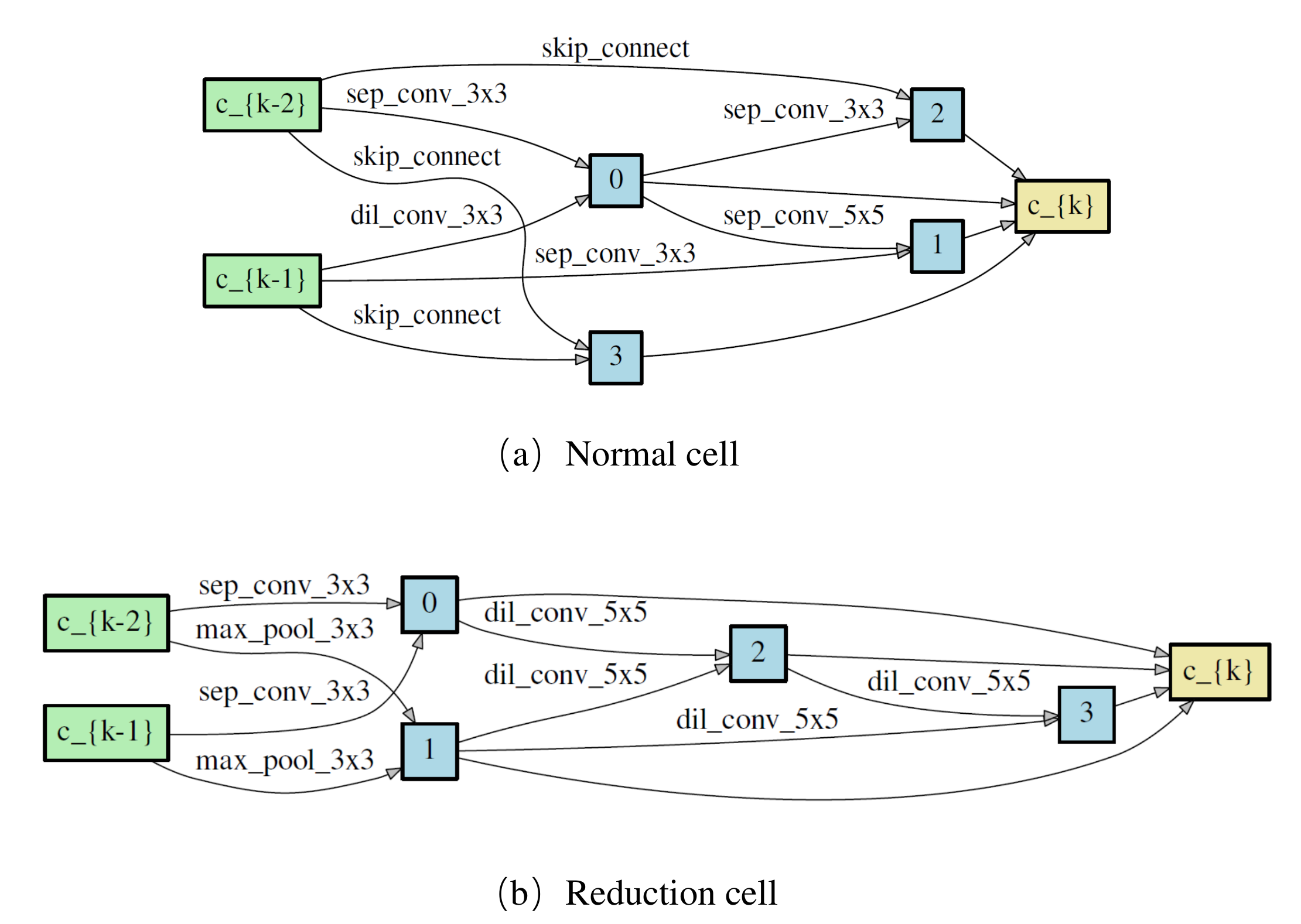}
\end{center}
\caption{Detailed structure of the best cells discovered on CIFAR-10 only by gradual operation pruning strategy. When pruning inferior operations and derive the final architecture during the search process, the operation importance are determined only by architecture weights.}
\label{fig:long}
\label{fig:onecol}
\end{figure}

Besides, compared with Figure 6 and Figure 7, we can see that EoiNAS encourages connections in a cell to cascade more levels, in other words, there are more layers in the cell, making the evaluation network further deeper and achieving better classification performance.

Finally, the combination of the operation importance indicator with the gradual operation pruning strategy can further enhance each other. The indicator is able to accurately represent the importance of operation and determine the remaining and pruning operations. Meanwhile, through gradually prune inferior operations, we can obtain more accurate indicator.

\section{Conclusion}

In this paper, we presented EoiNAS, a simple yet efficient architecture search algorithm for convolutional networks, in which a new indicator was proposed to fully exploit the operation importance to guide the model search. A gradual operation pruning strategy was proposed during the search process to further improve the search efficiency. By gradually pruning the inferior operations based on the proposed operation importance indicator, EoiNAS drastically reduced the computation consumption while achieving excellent model accuracies on CIFAR-10/100 and ImageNet, which outperformed the human-designed networks and other state-of-the-art NAS methods.

{\small
\bibliographystyle{ieee_fullname}
\bibliography{egbib}
}

\end{document}